\documentclass[10pt, conference, compsacconf]{IEEEtran}
\IEEEoverridecommandlockouts
\usepackage{cite}
\usepackage{amsmath,amssymb,amsfonts}
\usepackage{algorithm}
\usepackage{algorithmic}
\usepackage[pdftex]{graphicx}
\usepackage{textcomp}

\def\BibTeX{{\rm B\kern-.05em{\sc i\kern-.025em b}\kern-.08em
    T\kern-.1667em\lower.7ex\hbox{E}\kern-.125emX}}
\begin{document}

\title{Machine translation considering context information using Encoder-Decoder model\\}

\author{\IEEEauthorblockN{1\textsuperscript{st} Tetsuto Takano}
    \IEEEauthorblockA{
        \textit{Department of Electronic Information Science,}\\ 
        \textit{Kanazawa University,} \\
        \textit{Kanazawa, Japan,}\\
        \textit{Email: takanotetuto@gmail.com}
    }
    \and
    \IEEEauthorblockN{  2\textsuperscript{nd} Satoshi Yamane}
    \IEEEauthorblockA{
        \textit{Department of Electronic information Science,}\\ 
        \textit{Kanazawa University,} \\
        \textit{Kanazawa, Japan,}\\
        \textit{Email: syamane@is.t.kanazawa-u.ac.jp}
    }
}

\maketitle

\begin{abstract}
In the task of machine translation, context information is one of the important factor. 
But considering the context information model dose not proposed.
The paper propose a new model which can integrate context information and make translation.
In this paper, we create a new model based Encoder Decoder model.
When translating current sentence, the model integrates output from preceding encoder with current encoder. 
The model can consider context information and the result score is higher than existing model.
\end{abstract}

\section{Introduction}
In the field of natural language translation, using deep learning model gets great results. 
Among them, Encoder Decoder model [3] that proposed in 2014 captures the whole sentence as information gave great results.
The paper said its performance was reduced by 60\% as error rate compared with phrase-based machine translation.
\par
Current translation model makes sequence to sequence translation. 
In other words, the model outputs 1 translated sentence against 1 input sentence. 
However, we usually make translation considering the relationship with the preceding and following sentences. 
Therefore, it is difficult that the current translation model successfully translates sentences that are not completed by oneself. 
For example, it is assumed that a meaning of a polysemic word is mistaken . 
Therefore, a model which can not capture the context can not generate a correct translation and it is necessary to improve.
\par
The main purpose of this paper is to propose a new model that create translations considering the context of the preceding sentences.

\section{Related Work}
Recurrent Neural network Language model(RNNLM) [1], Long short terms memory (LSTM) [2] have high results in natural language processing using deep learning.
\par

\par
RNNLM is an application of RNN to a language model, which accepts variable length input by creating a loop in the network. 
However, RNN often cause a gradient explosion, so it is difficult to hold longterm memory.
\par
LSTM solve this problem. 
LSTM solve weak points of RNN by managing data to be stored by the gate.
With this characteristic, LSTM can hold longterm memory.

\section{PROPOSED TECHNIQUE}
\subsection{Model}
We use the Encoder Decoder model as the basis of our model. 
An example of the Encoder Decoder model is shown in Figure 1. 
In Figure 1, x and y represent word IDs, i and j represent word vectors, and p and q represent internal states of Encoder and Decoder.
t represents each state at time t.
\par
The Encoder Decoder model distributes learning objects by preparing LSTM for longterm memory in each of Encoder and Decoder, and achieves high results. 
On the Encoder side, the one-hot vector of the input word is converted into distributed representation, and LSTM reads it in order. 
After capturing end-of-sentence word, expresses the whole sentence as one vector and passes it to Decoder. 
On the Decoder side, after receiving the sentence vector, the translated result is generated referring to the received sentence vector and the word information which the model has generated in the past.
\par
In the model, Encoder outputs intermediate language of sentences [4]. 
An intermediate language is conceptual information of a sentence, the same expression is used in all languages, and conversion is not required.
\par
In this research, we aim to make translation with context information of the preceding sentence by improving Encoder Decoder model that output the intermediate language. 
As a feature of the model, it saves the Encoder output of the previous sentence and combines it with the output of Encoder of the translated sentence. 
Figure 2 shows proposed model. 
In Figure  2, $x$ and $y$ represent word IDs, $i$ and $j$ represent word vectors, and $p$ and $q$ represent internal states of Encoder and Decoder.
$\alpha_t^n$ represents each state at time $t$ in the $n$ th sentence.

\begin{figure}[t]
\centering
\includegraphics[keepaspectratio, scale=0.30]{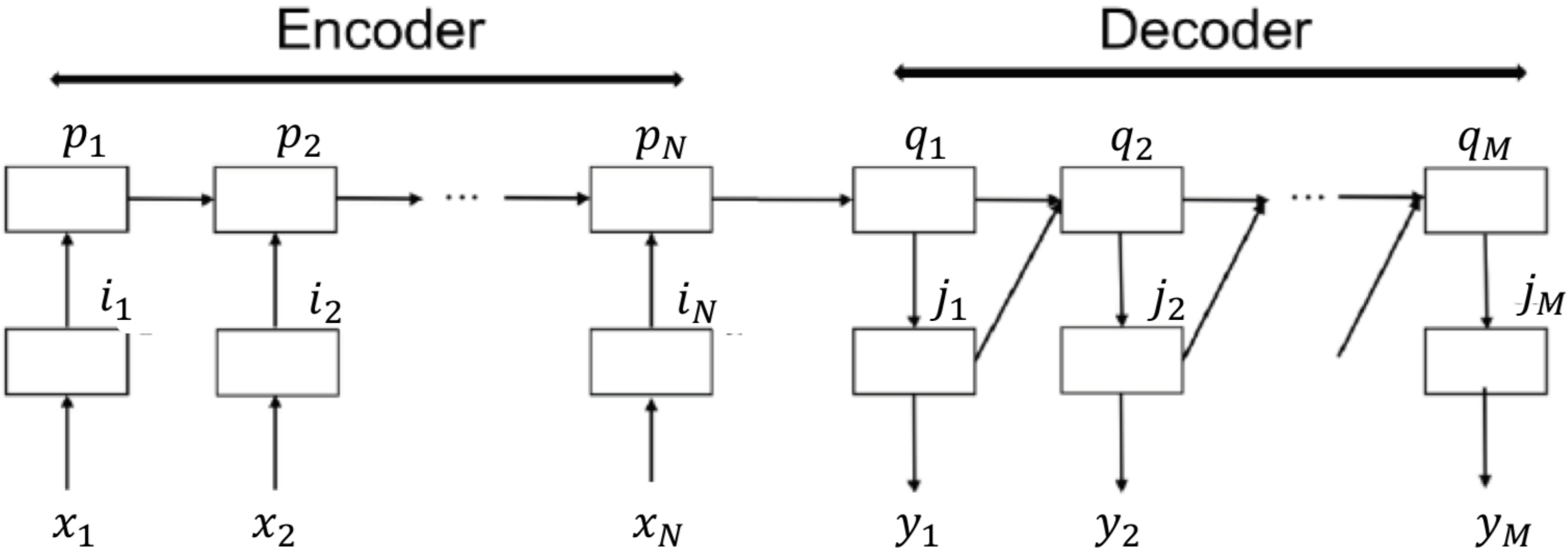}
\caption{Encoder Decoder model}
\label{fig}
\end{figure}

\begin{figure}[t]
\centering
\includegraphics[keepaspectratio, scale=0.30]{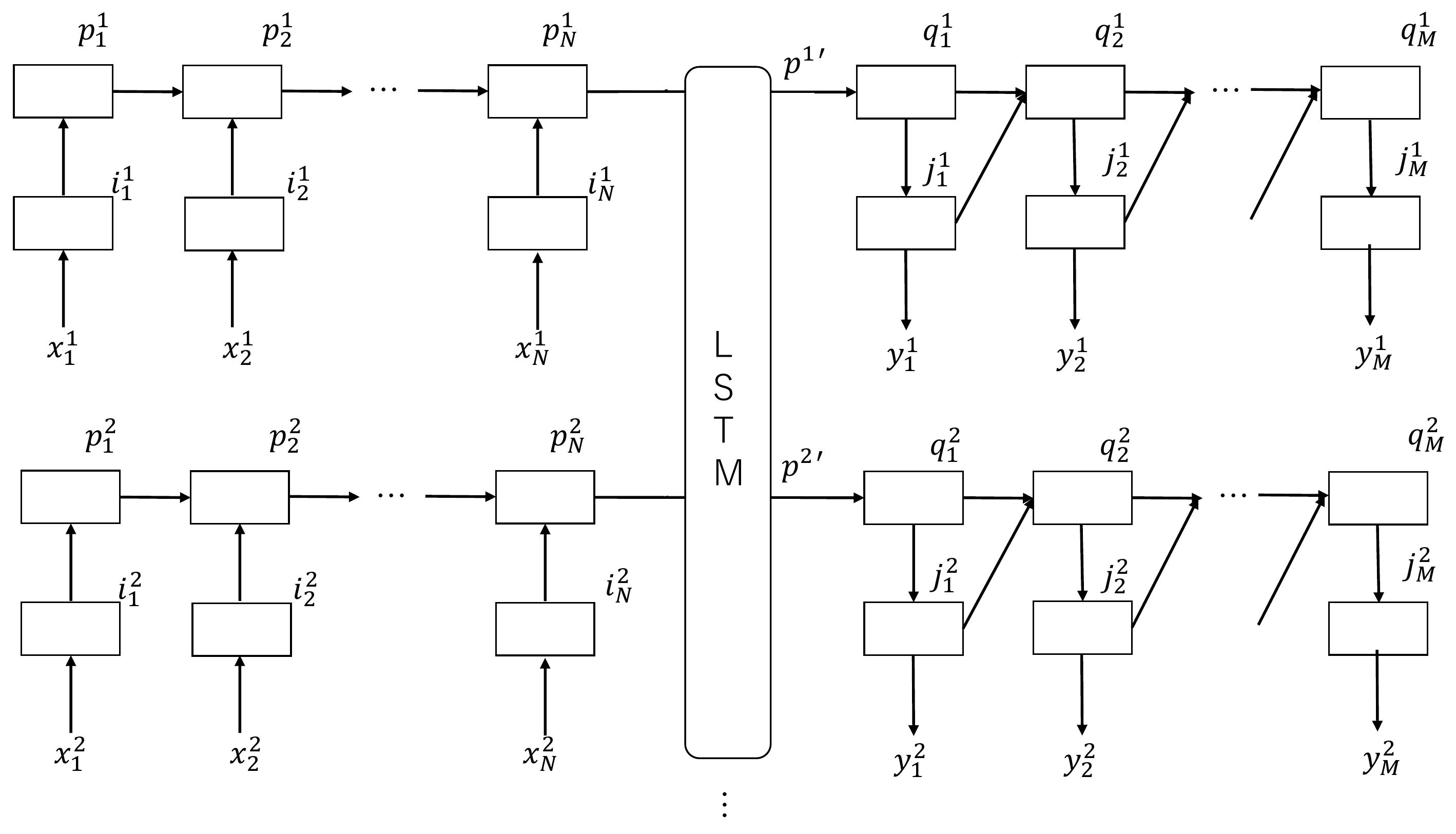}
\caption{proposed model}
\label{fig}
\end{figure}
    
\subsection{Algorithm}
We inserts vertical LSTM in Encoder Decoder model to incorporate not only horizontal word time series data but also vertical context time series data.
In the proposed model, if the current sentence is in the middle of a paragraph, the context information of the previous sentence is taken in and a translated sentence is output. 
If it is the beginning of a paragraph, the model translates only with the context information of the current sentence like the usual Encoder Decoder model.
\par
The algorithm is described in Algorithm 1.
\par
$E[l][n]$ is the word ID of the $n$ th source language with $n$ rows and $L$ and $N$ are the number of lines and the number of words in the source language.
\par
(1) and (5) respectively represent the conversion of one-hot expression and distributed expression of words,
(2) and (4) represent the internal state of the LSTM at the times $m$ and $n$, 
(6) represents the probability distribution of the next word. 
Compared to the Encoder Decoder model, a new point is (3). 
In line 10, it distinguish whichever the sentence is first sentence of context. And if so, $t$ th Encoder's output is combined with $t-1$ th Encoder's output.

\begin{algorithm}   
\caption{}                            
\label{alg1}                          
\begin{algorithmic}[1]    
\WHILE{$l < L$}
\WHILE{$n < N$} 
\STATE //Encoder prosess
\STATE{$n \Leftarrow  n+1$}
\STATE $x_n \Leftarrow E[l][n]$
\STATE $i_n \Leftarrow tanh(W_{xi}*x_n) $ (1)
\STATE $p_n \Leftarrow LSTM_{Encoder}(W_{ip}*i_n+W_{pp}*p_{n-1}) $ (2)
\ENDWHILE

\STATE //Connect prosess 
\IF{$E[l][0] == "**"$}
\STATE $q_1 \Leftarrow LSTM_{Connect}(W_{pq}*p_N)$
\ELSE 
\STATE $q_1 \Leftarrow LSTM_{Connect}(W_{pp\prime}*p_n+W_{p\prime p}*p_{n-1})$ (3)
\ENDIF
\STATE{$m \Leftarrow  1$}

\STATE //Decoder prosess
\REPEAT 
\STATE{$m \Leftarrow  m+1$}
\STATE{$q_m \Leftarrow LSTM_{Decoder}(W_{yq}*y_{m-1}+W_{qq}*q_{m-1}) $} (4)
\STATE{$j_m \Leftarrow tanh(W_{qj}*q_m)$} (5)
\STATE{$y_m \Leftarrow softmax(W_jy+j_m)$} (6)
\UNTIL{$y_m == EOS$}
\ENDWHILE
\end{algorithmic}
\end{algorithm}

\subsection{Dataset}
We use {\it Wikipedia Japanese-English Kyoto Related Documents Bilingual Corpus}\footnote{https://alaginrc.nict.go.jp/WikiCorpus/} which has 500,000 sentences of Japanese-English bilingual data as a dataset.
We consider the break of the paragraph as the break of the context. So we insert the word "*" to represent the sentence is first sentence of context. In addition, we replace number to 0 because number is easy to become unknown word by limitation of vocabulary. And if a word become unknown word, we replaced the word to "UNK".
\par
In this paper, we use 427,910 Japanese-English sentences as a learning data, and 21,456 sentences as a test data.

\subsection{Experiment and Evaluation}
The result is shown as Figure 3 and Figure 4. The orange line represents existing model's output and the blue line represents proposed model's.
In Figure 3, 1 iteration on the horizontal axis indicates 1 batch (64 sentences) was processed, and the vertical axis represents loss.
In Figure 4, 1 epoch on the horizontal axis indicates whole learning data was processed, and the vertical axis represents epoch.

In addition, We select two group of sentences from output of proposed model.
One is successfully incorporated a context information and generated a translation and the other is fail to translation.
Figure 5 shows successful one and Figure 6 shows unsuccessful one.

\subsection{Loss and BLEU}
Figure 3 and Figure 4 show that the proposed method succeeds in achieving higher results than the conventional method if same learning data are used.
Moreover, while the number of times of learning is small, the accuracy of the existing method is better, but the accuracy of the proposed method gradually improves as learning progresses. 
This is because the proposed method learns passing context information and normal translation at the same time, and can not generate a translation well if there are many learning objects and a small number of learnings, but as learning progressed, passing context information and translation can be learned well.
In fact, the BLEU score reaches a peak at around 50 epoch in the existing method, but in the proposed method, the score continues to grow smoothly without becoming a peak, and the score of the conventional method is over 100 epoch.

\begin{figure}[tbp]
\begin{center}
\includegraphics[width=80mm]{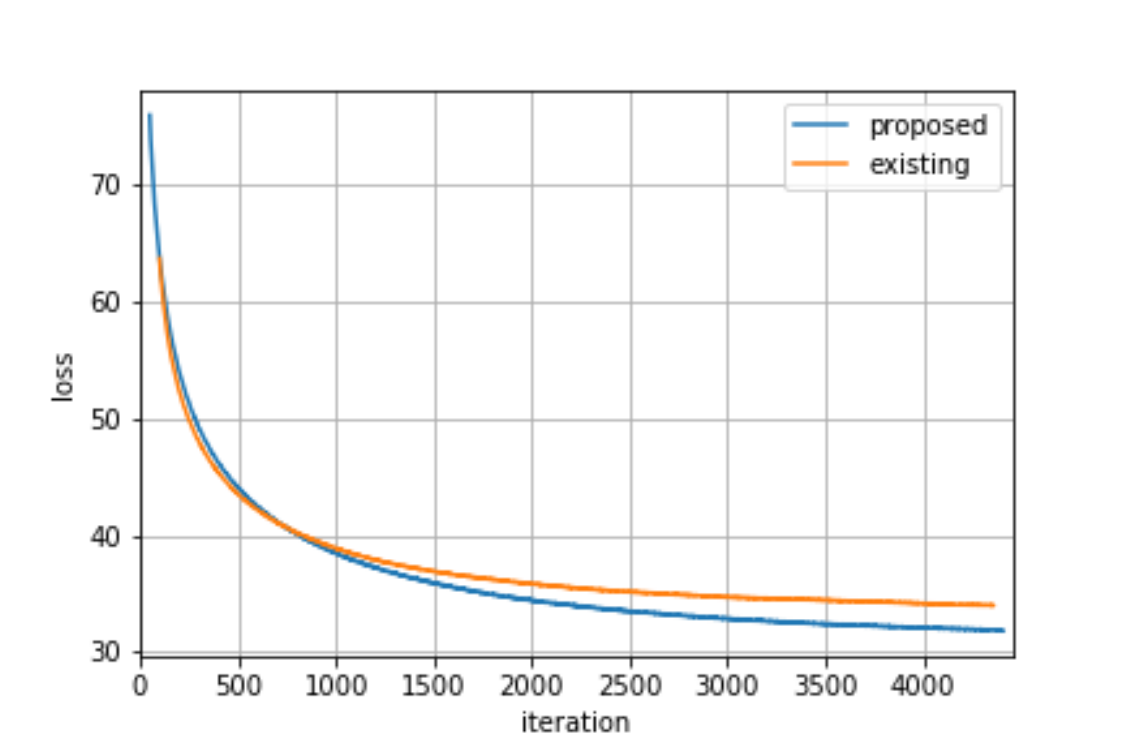}
\end{center}
\caption{1iteration(1batch)毎のloss}
\label{fig:gpuarch}
\end{figure}

\begin{figure}[tbp]
\begin{center}
\includegraphics[width=80mm]{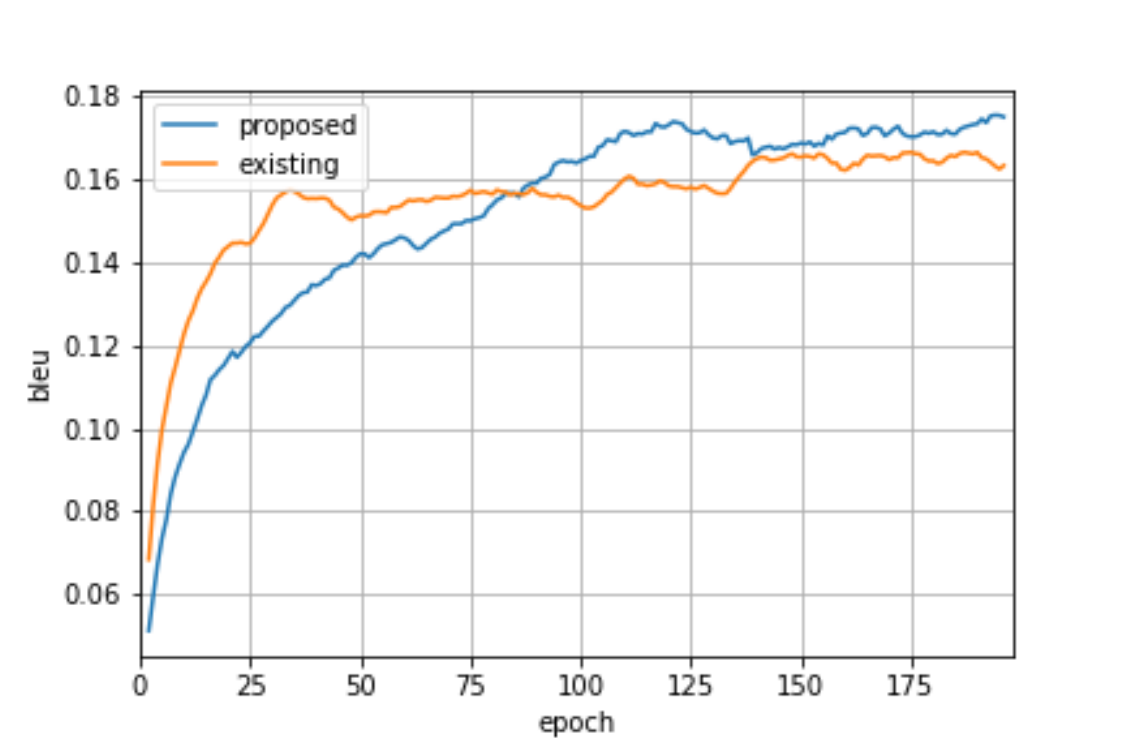}
\end{center}
\caption{BLEU}
\label{fig:gpuarch}
\end{figure}

\subsection{About output}
In the translation result of Figure 5, the context information up to the preceding sentence is well integrated, and the source language starting with "However" is regarded as reverse connection. However, in Figure 6, the translation is not successfully generated as compared with the conventional method. 
This is because the word "then", which is likely to be related to the previous sentence, takes a large proportion of the hidden layer of the previous sentence when mixing the output of the previous sentence's Encoder. 
So, the model failed to convey the state of the hidden layer with the information needed to translate.

\begin{figure}[tbp]
\begin{center}
\includegraphics[width=85mm]{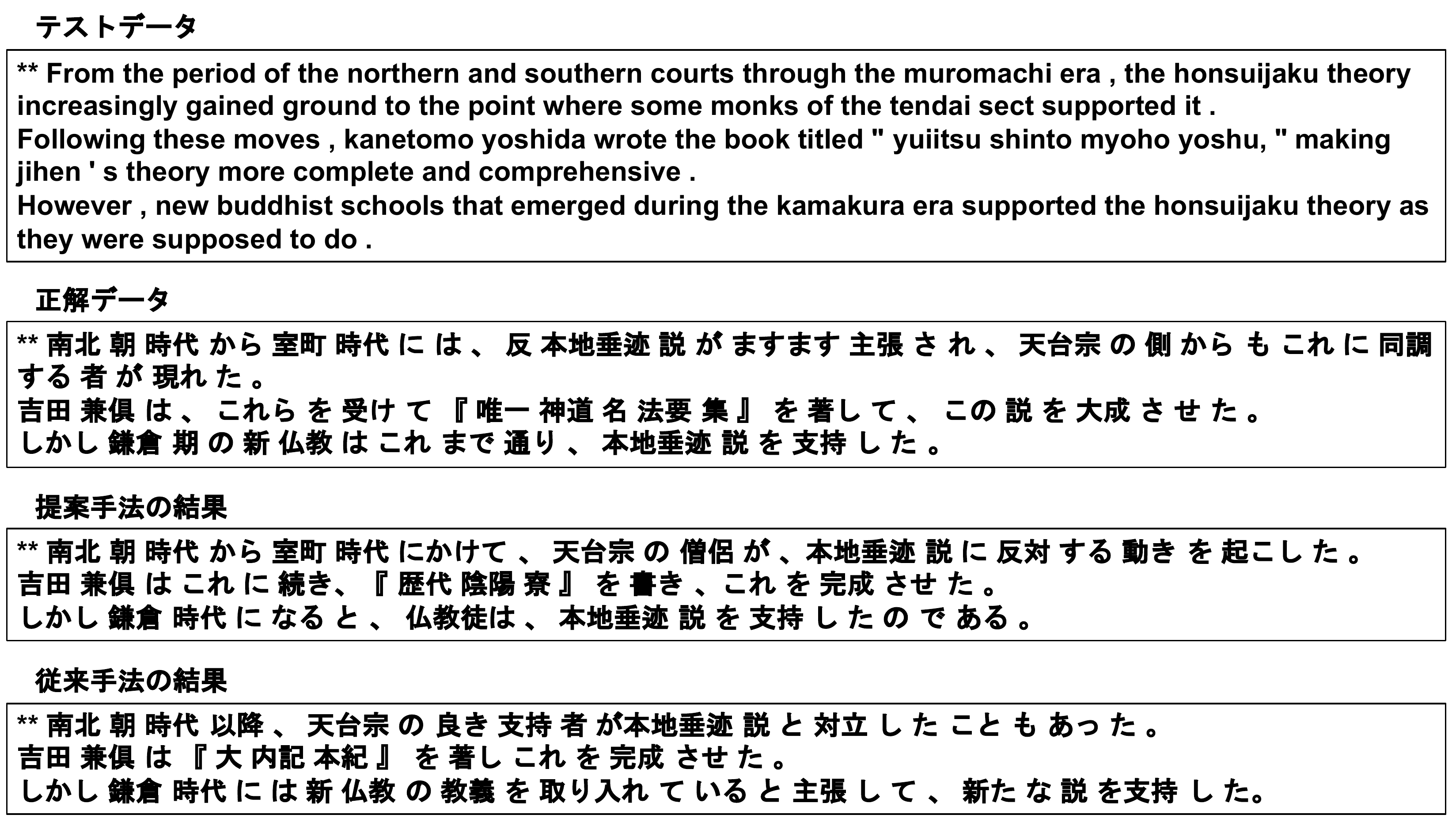}
\end{center}
\caption{output1 example of successful output}
\label{fig:gpuarch}
\end{figure}

\begin{figure}[tbp]
\begin{center}
\includegraphics[width=85mm]{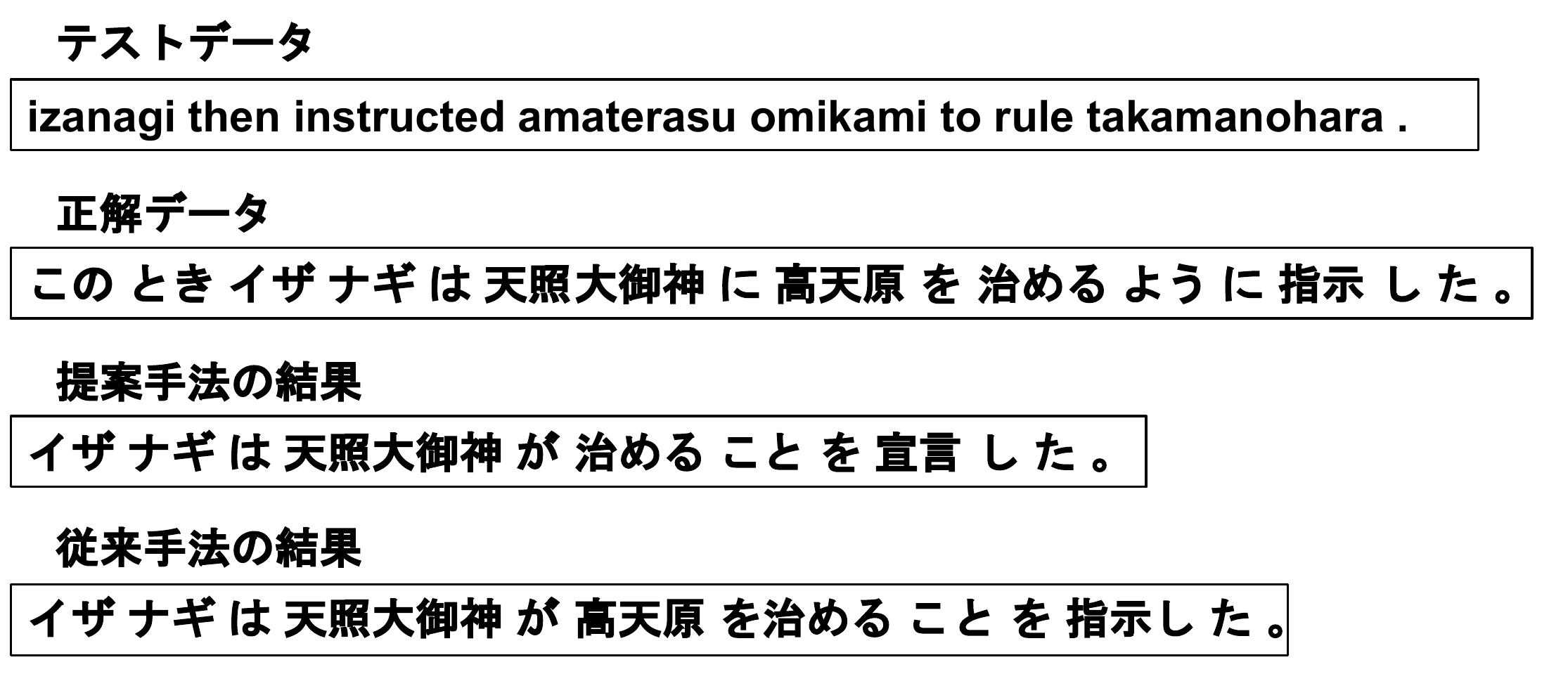}
\end{center}
\caption{output2 example of unsuccessful output}
\label{fig:gpuarch}
\end{figure}

\section{CONCLUSION}
In this research, we focused on the problem of not being able to capture context information of the existing Encoder Decoder model, and proposed a model that could capture context information. 
Moreover, when we do experiment with the model, the accuracy was inferior to the conventional method in a small number of learnings, but when learning progressed, the result that exceeds the conventional method was obtained. 
When we compared the translation results of the two models, it was found that the proposed method understands the context and generates a translation. 
As a result, we succeeded in further extending the translation accuracy that gets stuck in the middle of learning in existing model by incorporating context information
\par
The following two points will be mentioned as future issues. 
The first is the lack of data sets. 
In this paper, We do an experiment only with 1/500 of the data set of the conventional method. 
In research using deep learning, the size of the learning data is directly linked to the accuracy, and this is a problem that must be resolved as soon as possible.
\par
Second, the context information of the sentence currently being translated may not be transmitted well and the information may be lost if too much context information of the previous sentence is incorporated.
In most cases, the most important part of the information needed to translate a sentence is considered to be the translation sentence.
So, it may be necessary to intentionally limit the context information up to the preceding sentence.
\par


\begin{thebibliography}{00}
\bibitem{b2} J. L. Elman, ''Finding structure in time,'' Cognitive Science Volume 14, Issue 2, pp. 179-211,1990.
\bibitem{b3} S. Hochreiter and J. Schmidhuber, ''Long shortterm memory,'' Neural computation Vol. 9,No. 8, pp.1735-1780, Nov. 1997.
\bibitem{b4} I. Sutskever, O. Vinyals and Q. V. Le, ''Sequence to sequence learning with neural networks,'' Proceedings of Advances in Neural Information Processing Systems, pp. 3104-3112, Dec. 2014.
\bibitem{b5} Melvin Johnson , Mike Schuster, Quoc V. Le, Maxim Krikun, Yonghui Wu, Zhifeng Chen, Nikhil Thorat, Fernanda Vigas, ''Google’s Multilingual Neural Machine Translation System: Enabling Zero-Shot Translation'' Nov 2016
\end{thebibliography}
\end{document}